\documentclass{article}

\usepackage{microtype}
\usepackage{graphicx}
\usepackage{subcaption}
\usepackage{booktabs} 

\usepackage{hyperref}

\usepackage[preprint]{icml2026}

\usepackage{amsmath}
\usepackage{amssymb}
\usepackage{mathtools}
\usepackage{amsthm}

\definecolor{sftcolor}{RGB}{220, 120, 90}
\definecolor{rftcolor}{RGB}{1, 87, 147}
\usepackage[table]{xcolor}  
\usepackage{booktabs}       
\usepackage{tcolorbox}      
\tcbuselibrary{skins, breakable} 

\newcounter{globalpromptcount}

\newtcolorbox{promptbox}[2]{
    colback=white,
    colframe=rftcolor!80,
    arc=2pt,
    outer arc=2pt,
    breakable,
    title={%
        \refstepcounter{globalpromptcount}%
        \textbf{Prompt~\theglobalpromptcount:} #1%
        \label{#2}%
    }
}
\graphicspath{{./figs/}}
\newcommand{\coloredunderbrace}[3]{%
  {\color{#1}\underbrace{\color{black}#2}_{\textcolor{#1}{\text{#3}}}}%
}
\usepackage{multirow}
\usepackage{pifont}
\usepackage[ruled,vlined,algo2e]{algorithm2e}
\newcommand{\ours}{ViSurf} 
\usepackage{enumitem}

\usepackage[capitalize,noabbrev]{cleveref}

\theoremstyle{plain}

\theoremstyle{definition}

\theoremstyle{remark}

\usepackage[textsize=tiny]{todonotes}

\icmltitlerunning{{\ours}: Visual Supervised-and-Reinforcement Fine-Tuning}

\begin{document}

\twocolumn[
  \icmltitle{{\ours}: Visual Supervised-and-Reinforcement Fine-Tuning for Large Vision-and-Language Models}



  \icmlsetsymbol{equal}{*}

    \begin{icmlauthorlist}
    \icmlauthor{Yuqi Liu}{cuhk}
    \icmlauthor{Liangyu Chen}{ruc}
    \icmlauthor{Jiazhen Liu}{hkust}
    \icmlauthor{Mingkang Zhu}{cuhk}
    \icmlauthor{Zhisheng Zhong}{cuhk}
    \icmlauthor{Bei Yu}{cuhk}
    \icmlauthor{Jiaya Jia}{hkust}

  \end{icmlauthorlist}

  \icmlaffiliation{cuhk}{The Chinese University of Hong Kong}
  \icmlaffiliation{ruc}{Renmin University of China}
  \icmlaffiliation{hkust}{The Hong Kong University of Science and Technology}

  \icmlcorrespondingauthor{}{}

  \icmlkeywords{Machine Learning, ICML}

  \vskip 0.3in
]



\printAffiliationsAndNotice{}  

\begin{abstract}
Post-training Large Vision-and-Language Models (LVLMs) typically involves Supervised Fine-Tuning (SFT) for knowledge injection or Reinforcement Learning with Verifiable Rewards (RLVR) for performance enhancement. However, SFT often leads to sub-optimal performance, while RLVR remains constrained by the model's internal knowledge base. While a sequential SFT $\rightarrow$ RLVR pipeline can be used, it introduces significant computational overhead and suffers from catastrophic forgetting. To address these limitations, we propose ViSurf (\textbf{Vi}sual \textbf{Su}pervised-and-\textbf{R}einforcement \textbf{F}ine-Tuning), a unified, single-stage paradigm that integrates the strengths of both SFT and RLVR. By analyzing their training objectives, we establish a unified framework that injects ground-truth labels directly into RLVR rollouts, facilitating simultaneous external supervision and internal reinforcement. Furthermore, we introduce three novel reward control strategies to ensure training stability and optimization. Extensive experiments demonstrate that ViSurf consistently outperforms standalone SFT, RLVR, and the traditional two-stage pipeline across diverse benchmarks. In-depth analysis corroborates these findings, validating the derivation and design principles of ViSurf. 
\end{abstract}



\section{Introduction}
\label{sec:intro}

\begin{figure}
    \centering
    \includegraphics[width=0.95\linewidth]{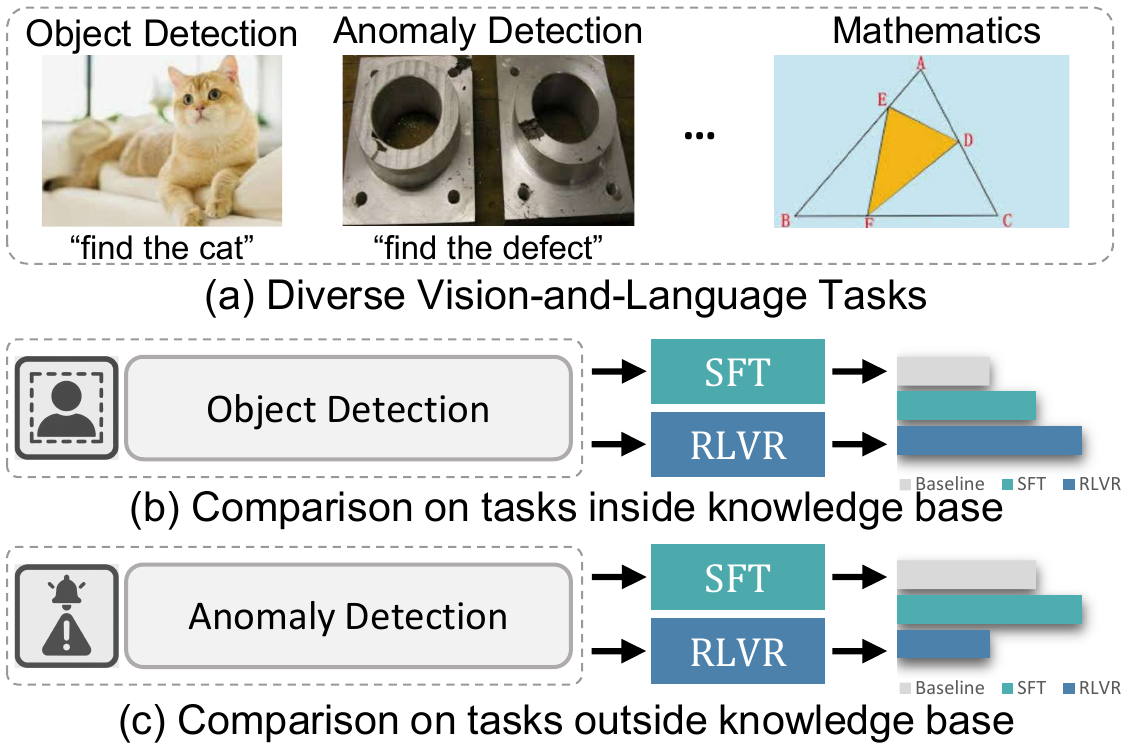}
    \caption{(a) Examples of vision-ang-language tasks. (b) For tasks within LVLMs' knowledge base, RLVR performs better than SFT. (c) For tasks that exceed LVLMs' knowledge, SFT performs better, whereas RLVR performs worse than baseline.}
    \label{fig:teaser}
\end{figure}

Developing Large Vision-and-Language Models (LVLMs) capable of excelling across diverse visual perception tasks represents a pivotal frontier in visual intelligence. To date, research has predominantly relied on two distinct post-training paradigms: Supervised Fine-Tuning (SFT) \citep{liu2023visual,wang2024qwen2vl,bai2025qwen25vl} and Reinforcement Learning with Verifiable Rewards (RLVR) \citep{liu2025seg,liu2025visionreasoner}. 

However, these paradigms exhibit a fundamental dichotomy in their strengths and weaknesses. SFT directly optimizes models using expert-annotated data, providing explicit external guidance that enables the model to approximate target distributions. Despite this, SFT often yields sub-optimal generalization and is prone to catastrophic forgetting of pre-trained knowledge. Conversely, RLVR, particularly on-policy methods like Group Relative Policy Optimization (GRPO) \citep{shao2024deepseekmath} and Dynamic Sampling Policy Optimization (DAPO) \citep{yu2025dapo}, leverages internal reinforcement signals to optimize policies based on rollout evaluations. While RLVR effectively mitigates forgetting and enhances performance, its efficacy is fundamentally capped by the model's inherent knowledge boundaries; performance typically stagnates when tasks extend beyond the initial policy's distribution. 
Our evaluation across diverse vision-language benchmarks, summarized in \Cref{fig:teaser}, confirms this phenomenon: SFT excels in out-of-distribution knowledge acquisition, whereas RLVR thrives on tasks aligning with pre-existing capabilities (see \Cref{sec:case_study} for a detailed case study). While a sequential SFT $\rightarrow$ RLVR pipeline attempts to marry these strengths, it introduces prohibitive computational overhead and remains vulnerable to forgetting during the initial SFT phase.

To overcome these challenges, we propose ViSurf (\textbf{Vi}sual \textbf{Su}pervised-and-\textbf{R}einforcement \textbf{F}ine-Tuning), a unified, single-stage paradigm that integrates the complementary advantages of SFT and RLVR. We provide a rigorous analysis of the underlying objectives and gradients for both methods, theoretically demonstrating that their shared gradient patterns allow for integration into a singular ViSurf objective. Unlike existing approaches \citep{zhang2025policy,ma2025learning} that simply sum the SFT and RLVR losses, ViSurf offers a theoretically grounded, unified perspective. 
While subtle differences exist, the gradient of {\ours} objective can be interpreted as a composite of the gradients from both SFT and RLVR. 
To ensure training stability within this unified framework, we introduce three novel reward control strategies for ground-truth labels: (i) preference alignment with policy rollouts, (ii) exclusion of "thinking" rewards for static labels, and (iii) reward smoothing to prevent optimization spikes. Consequently, the implementation of ViSurf is streamlined into an elegant process: interleaving ground-truth demonstrations with on-policy rollouts within a single, unified training phase.

\begin{figure}
    \centering
    \includegraphics[width=1.\linewidth]{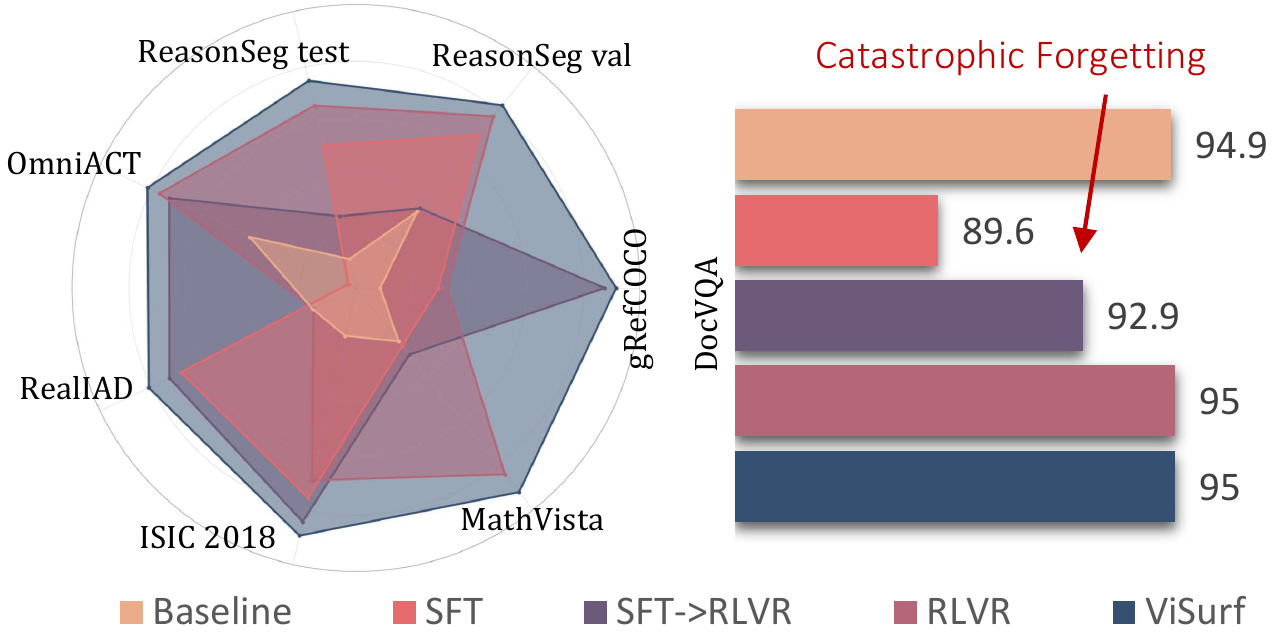}
    \caption{Radar Chart: {\ours} achieves superior performance across different training paradigms. Bar Chart: SFT and two-stage SFT \textrightarrow ~RLVR exhibit catastrophic forgetting.}
    \label{fig:teaser_performance}
\end{figure}

We evaluate {\ours} across diverse domains, with a comparative summary presented in \Cref{fig:teaser_performance}. The results demonstrate that {\ours} consistently outperforms SFT, RLVR, and the SFT $\rightarrow$ RLVR pipeline. Notably, our method mitigates catastrophic forgetting, as evidenced by its stable performance on VQA benchmarks. Furthermore, our ablation study confirms the critical role of the proposed reward control mechanism, while in-depth analysis provides empirical support for our theoretical framework and offers insights into the operational principles of {\ours}. Our contributions are summarized as follows:

\begin{itemize}[itemsep=1pt, topsep=1pt]
    \item \textbf{Unified Framework}: We introduce {\ours}, a theoretically grounded, single-stage paradigm that unifies SFT and RLVR to achieve simultaneous knowledge injection and internal reinforcement.
    \item \textbf{Stability Optimization}: We design three novel reward control strategies to stabilize joint optimization, effectively balancing ground-truth guidance with on-policy exploration. 
    \item \textbf{State-of-the-Art Performance}: {\ours} consistently outperforms standalone SFT, RLVR, and two-stage pipelines across diverse benchmarks, supported by in-depth analysis of its synergistic mechanics.
\end{itemize}

\section{Related Works}
\label{sec:related_works}

\subsection{Supervised Fine-tuning for LVLMs}
Supervised Fine-Tuning (SFT) has emerged as the predominant paradigm for advancing Large Vision-Language Models (LVLMs), typically involving the refinement of pre-trained models using expert-annotated datasets. Prominent examples include the LLaVA-series \citep{liu2023visual,li2024llavaov,liu2024improved,an2025llava}, QwenVL-series \citep{wang2024qwen2vl,bai2025qwen25vl}, MGM-series \citep{li2024mini,zhong2024lyra,wang2025mgm}, Eagle \citep{chen2025eagle} and InternVL-series \citep{chen2024internvl,zhu2025internvl3}, all of which leverage the SFT methodology. Beyond foundational training, SFT has demonstrated remarkable efficacy in tailoring LVLMs to specialized downstream tasks, such as image quality assessment \citep{you2025teaching} and autonomous driving \citep{xu2025drivegpt4}.

\subsection{Reinforcement Learning for LVLMs}
Beyond supervised methods, Reinforcement Learning (RL) has become a cornerstone for aligning Large Vision-Language Models (LVLMs) with human expectations. Established techniques like DPO \citep{rafailov2023direct} necessitate extensive human-labeled preference datasets, which are often prohibitively expensive to curate. Similarly, PPO \citep{schulman2017proximal} relies on a robust reward model to supervise the policy’s outputs. To mitigate these constraints, Reinforcement Learning from Verifiable Rewards (RLVR) algorithms, such as GRPO \citep{shao2024deepseekmath} and DAPO \cite{yu2025dapo}, leverage verification functions to evaluate model performance. The potency of this paradigm for LVLMs is underscored by recent advancements \citep{liu2025seg,liu2025visual,liu2025visionreasoner,liu2025empowering,huang2025vision}, most notably in SegZero \citep{liu2025seg} and VisualRFT \cite{liu2025visual}.

\section{{\ours}}

We begin by analyzing the objective functions of SFT and RLVR in \Cref{sec:preliminary}. A case study in \Cref{sec:case_study} then highlights the limitations of both methods. To address these limitations, we introduce {\ours}, detailing its design and its theoretical alignment with SFT and RLVR in \Cref{sec:combine_sft_rft}. Subsequently, \Cref{sec:reward_control} presents three novel mechanisms for reward control during training, and \Cref{sec:optimization_analysis} a rigorous analysis of the underlying optimization dynamics.

\subsection{Preliminary}
\label{sec:preliminary}

Let $\pi_{\theta}$ denote a large vision-and-language model (LVLM), parameterized by $\theta$. Common post-training paradigms for optimizing $\pi_{\theta}$ include Supervised Fine-Tuning (SFT) and Reinforcement Learning with Verifiable Rewards (RLVR). Both SFT and RLVR utilize the same input dataset, $\mathcal{D}_{\text{input}}=\{{(v_i, t_i)\}}^N_{i=1}$, where $v_i$ is a visual input, $t_i$ is a textual input, and $N$ is the dataset size.

\textbf{Supervised Fine-Tuning (SFT)} optimizes $\pi_{\theta}$ against a set of ground-truth labels, $\mathcal{D}_{\text{label}}=\{y_i\}^N_{i=1}$.
The objective is to minimize the negative log-likelihood of the labels:

\begin{equation}
\label{eq:sft}
    \mathcal{L}_{\text{SFT}} (\theta) 
    = - \mathbb{E}_{\substack{(v,t) \sim \mathcal{D}_{\text{input}} \\ \textcolor{sftcolor}{y \sim \mathcal{D}_{\text{label}}}}} 
    \left[ \log \pi_{\theta}(y \mid v,t) \right],
\end{equation}

\noindent where $y$ corresponds to $(v,t)$. A more precise notation would be $(v,t,\textcolor{sftcolor}{y}) \sim \text{zip} (\mathcal{D}_{\text{input}}, \textcolor{sftcolor}{\mathcal{D}_{\text{label}}})$. 
Nevertheless, we retain the current notation, $(v,t) \sim \mathcal{D}_{\text{input}}, \textcolor{sftcolor}{y \sim \mathcal{D}_{\text{label}}}$, for clarity and ease of comparison in the subsequent discussion.

\textbf{Reinforcement Learning with Verifiable Rewards (RLVR).} We illustrate RLVR using the on-policy Group Relative Policy Optimization (GRPO) algorithm \citep{shao2024deepseekmath}. GRPO optimizes the policy $\pi_{\theta}$ using a verifiable reward function, which typically combines measures of output format and accuracy \cite{guo2025deepseek,liu2025visionreasoner,liu2025seg}. 
For a given input $(v_i,t_i) \in \mathcal{D}_{\text{input}}$, the old policy $\pi_{\theta_{old}}$ (from a previous optimization step) generates a group of $G$ rollouts $\{o_{j}\}^G_{j=1}$ by sampling with different random seeds. Each rollout $o_j$ is then evaluated by a reward function $r(\cdot)$, resulting in a set of rewards $\{r(o_{j})\}^G_{j=1}$. The advantage for each rollout is subsequently computed as follows:


\begin{equation}
\footnotesize
\label{eq:adv}
    \hat{A}_j = \frac{\text{r}(o_{j}) - \text{mean}\left(\{\text{r}(o_{j})\}^G_{j=1}\right)}
    {\text{std}\left(\{\text{r}(o_{j})\}^G_{j=1}\right)},
\end{equation}


\noindent The objective of RLVR is to minimize the equation:

\begin{equation}
\footnotesize
\label{eq:rft}
\begin{aligned}
\mathcal{L}_{\mathrm{RLVR}}(\theta)
&= -\mathbb{E}_{\substack{(v,t)\sim\mathcal{D}_{\mathrm{input}} \\ \textcolor{rftcolor}{ \{o_j\}_{j=1}^G \sim \pi_{\theta_{\mathrm{old}}}}}}
      \left[ \frac{1}{G} \sum_{j=1}^G
      \min \Bigg\{
        \frac{\pi_\theta(o_j \mid v,t)}{\pi_{\theta_{\mathrm{old}}}(o_j \mid v,t)} \,\textcolor{rftcolor}{\hat{A}_j}, \right. \\[6pt]
&\qquad\qquad \left.
        \operatorname{clip}\!\Biggl(
          \frac{\pi_\theta(o_j \mid v,t)}{\pi_{\theta_{\mathrm{old}}}(o_j \mid v,t)},
          \;1-\epsilon,\;1+\epsilon
        \Biggr)\,\textcolor{rftcolor}{\hat{A}_j}
      \Bigg\}
      \right],
\end{aligned}
\end{equation}

\noindent where the $\epsilon$ is a constant that controls the clipping boundary. For simplicity, both in equation and in our practical implementation, we remove the KL divergence term.

\subsection{Case Study: Non-Object Scenarios}
\label{sec:case_study}

\begin{figure}
    \centering
    \includegraphics[width=1.\linewidth]{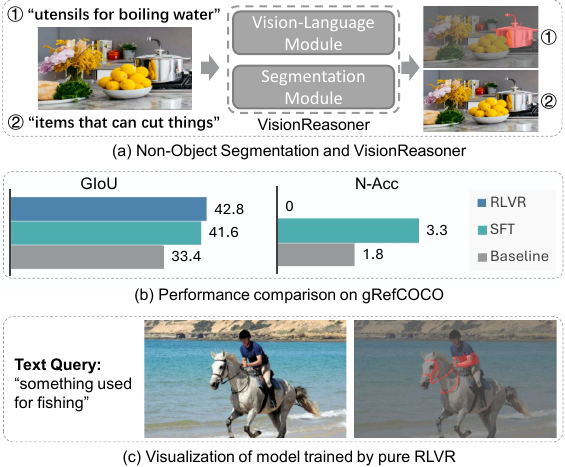}
    \caption{(a) Illustration on Non-Object Segmentation and VisionReasoner. (b) Performance comparison of SFT and RLVR on gRefCOCO. While RLVR achieves higher overall GIoU, it fails on non-object instructions. (c) Specifically, the RLVR model consistently outputs a mask even when no relevant object is detected.}
    \label{fig:performance_comparison}
\end{figure}

We conduct a case study on non-object referring expression segmentation, a challenging task where instructions include both valid expressions and ``incorrect" ones referring to non-existent objects. 
For this study, we utilize the VisionReasoner architecture \cite{liu2025visionreasoner}, which is initialized with Qwen2.5-VL \citep{bai2025qwen25vl} and SAM2 \citep{ravi2024sam2} and has demonstrated strong performance on standard referring segmentation tasks. 
\Cref{fig:performance_comparison}(a) illustrates the non-object segmentation and the architecture of VisionReasoner. Experimental configurations are detailed in \Cref{sec:expri_setting}.

The gRefCOCO \cite{liu2023gres} serves as the benchmark for this scenario. We compare gIoU (average IoU across images) and N-Acc (accuracy in identifying non-existent object).
Quantitative and qualitative results are presented in \Cref{fig:performance_comparison}(b)-(c). Our analysis reveals that model trained by SFT achieves suboptimal gIoU performance but tends to learn to correctly identify the absence of objects. In contrast, model trained with RLVR attain higher overall gIoU scores, yet they often generate object masks even when no relevant object is present. This limitation arises because the RLVR model, relying solely on self-rollouts, lacks the corrective mechanism necessary to produce a correct “no object” output.
In essence, RLVR drives performance gains through self-exploration, whereas SFT provides the external grounding necessary when exploration fails. This dichotomy motivates our central research question: \textit{how can we efficiently synthesize the strengths of both training paradigms within a unified training stage?}

\subsection{Combining SFT and RLVR}
\label{sec:combine_sft_rft}

To harness the complementary benefits of SFT and RLVR, we propose {\ours}: a unified, single-stage post-training algorithm. This section elaborates on the derivation of the {\ours} algorithm and its relation to SFT and RLVR.

\textbf{Gradient Analysis of SFT and RLVR.} The gradient of SFT can be derived from \Cref{eq:sft} as: 

\begin{equation}
\footnotesize
\label{eq:delta_sft}
\nabla_{\theta}\mathcal{L}_{\mathrm{SFT}}(\theta)
= -\mathbb{E}_{\substack{(v,t)\sim\mathcal{D}_{\mathrm{input}} \\ \textcolor{sftcolor}{y\sim\mathcal{D}_{\mathrm{label}}}}}
\left[\nabla_{\theta}\log \pi_{\theta}(y \mid v,t)\right].
\end{equation}

\noindent In our practical implementation, we adopt a single-sampling, single-updating paradigm, which effectively bypasses the need for clipping operations. Therefore, the gradient of RLVR can be derived from \Cref{eq:rft} using approximation $\pi_{\theta} \approx \pi_{\theta_{old}}$ and log-derivative trick: 

\begin{equation}
\footnotesize
\label{eq:delta_rft}
\begin{aligned}
\nabla_{\theta}\mathcal{L}_{\mathrm{RLVR}}(\theta)
&= -\mathbb{E}_{\substack{(v,t)\sim\mathcal{D}_{\mathrm{input}}\\ 
     \textcolor{rftcolor}{ \{o_j\}_{j=1}^G\sim\pi_{\theta_{\mathrm{old}}}}}} \\[6pt]
&\quad \left[ \frac{1}{G}\sum_{j=1}^G 
      \textcolor{rftcolor}{\hat A_j}\,\nabla_{\theta}\log\pi_{\theta}(o_j\mid v,t) \right]_{\,\theta \approx \theta_{\mathrm{old}}}.
\end{aligned}
\end{equation}

We observe that the gradients of the SFT and RLVR losses,  $\nabla_{\theta}\mathcal{L}_{\mathrm{SFT}}(\theta)$ and $\nabla_{\theta}\mathcal{L}_{\mathrm{RLVR}}(\theta)$, share a similar form. The difference between them is the guidance signal ($y$ vs. $\{o_j\}_{j=1}^G$) and coefficient ($1$ vs. $\hat A_j$).

\textbf{Objective of {\ours}.} To combine SFT and RLVR into a single stage, we design an objective function that naturally yields a gradient combining both $\nabla_{\theta}\mathcal{L}_{\mathrm{SFT}}(\theta)$ and $\nabla_{\theta}\mathcal{L}_{\mathrm{RLVR}}(\theta)$. Our key insight is to include the ground-truth label $y$ as a high-reward sample within the RLVR framework. We construct an augmented rollout set $y \cup \{o_j\}_{j=1}^G$. Then the corresponding rewards are $\text{r}(y) \cup \{\text{r}(o_{j})\}^G_{j=1}$. This formulation modifies the advantage calculation of rollouts in \Cref{eq:adv} as follows:
\begin{equation}
\footnotesize
\label{eq:adv_rollouts}
    \hat{A}_j = \frac{\text{r}(o_{j}) - \text{mean}\left(\text{r}(y) \cup \{\text{r}(o_{j})\}^G_{j=1}\right)}
    {\text{std}\left(\{\text{r}(y) \cup \{\text{r}(o_{j})\}^G_{j=1}\}\right)},
\end{equation}

\noindent and the advantage of ground-truth $y$ is calculated as: 

\begin{equation}
\footnotesize
\label{eq:adv_y}
    \hat{A}_y = \frac{\text{r}(y) - \text{mean}\left(\text{r}(y) \cup \{\text{r}(o_{j})\}^G_{j=1}\right)}
    {\text{std}\left(\{\text{r}(y) \cup \{\text{r}(o_{j})\}^G_{j=1}\}\right)}.
\end{equation}

\noindent The objective of {\ours} is to minimize the equation:

\begin{equation}
\footnotesize
\label{eq:visual_srft}
\begin{aligned}
\mathcal{L}_{\mathrm{{\ours}}} & (\theta)
= -\mathbb{E}_{\substack{
      (v,t)\sim\mathcal{D}_{\mathrm{input}} \\
      \textcolor{rftcolor}{\{o_j\}_{j=1}^G \sim \pi_{\theta_{\mathrm{old}}}} \\
      \textcolor{sftcolor}{y \sim \mathcal{D}_{\text{label}}}}} \\[6pt]
&
      \Bigg[
        \frac{1}{G+1} \Bigg( 
              \sum_{j=1}^G \min \Bigg\{ 
              \frac{\pi_\theta(o_j \mid v,t)}{\pi_{\theta_{\mathrm{old}}}(o_j \mid v,t)}
              \,\textcolor{rftcolor}{\hat{A}_j}, \\[6pt]
&\quad
              \operatorname{clip}\!\Biggl(
                \frac{\pi_\theta(o_j \mid v,t)}{\pi_{\theta_{\mathrm{old}}}(o_j \mid v,t)},\;
                1-\epsilon,\;1+\epsilon
              \Biggr)\,\textcolor{rftcolor}{\hat{A}_j}
          \Bigg\} \\[8pt]
&
          + \min \Bigg\{
              \frac{\pi_\theta(y \mid v,t)}{\pi_{\theta_{\mathrm{old}}}(y \mid v,t)}
              \,\textcolor{sftcolor}{\hat{A}_y}, \\[6pt]
&\quad
              \operatorname{clip}\!\Biggl(
                \frac{\pi_\theta(y \mid v,t)}{\pi_{\theta_{\mathrm{old}}}(y \mid v,t)},\;
                1-\epsilon,\;1+\epsilon
              \Biggr)\,\textcolor{sftcolor}{\hat{A}_y}
          \Bigg\}
        \Bigg)
      \Bigg].
\end{aligned}
\end{equation}

\noindent With the objective function demonstrated above, the pseudocode of {\ours} Optimization Step is shown in \Cref{alg:visualsrft}.



\begin{algorithm2e}[t] 
\footnotesize
\caption{{\ours} Optimization Step}
\label{alg:visualsrft}
\DontPrintSemicolon 

\KwIn{
  \parbox[t]{0.85\linewidth}{
    policy model $\pi_{\theta}$; reward function $\text{r}(\cdot)$; \\
    input data $\mathcal{D}_{\mathrm{input}}$; label data $\mathcal{D}_{\mathrm{label}}$
  }
}

\For{step $=1, \ldots, M$}{
    Sample a mini-batch $\mathcal{B}_\mathrm{input}$ and corresponding $\mathcal{B}_{\mathrm{label}}$\;
    
    Update the old policy model $\pi_{\theta_{\text{old}}} \gets \pi_{\theta}$\;
    
    Sample $G$ outputs $\{o_j\}_{j=1}^G \sim \pi_{\theta_{\text{old}}}()$ \\
    \quad for each $(v,t) \in \mathcal{B}_\mathrm{input}$\;
    
    Compute rewards $\{\text{r}(o_j)\}_{j=1}^G$ for each sampled output $o_j$\;
    
    Compute rewards $\text{r}(y)$ for label $y \in \mathcal{B}_{\mathrm{label}}$\;
    
    Compute $\hat{A}_j$ and $\hat{A}_y$ through relative advantage estimation\;
    
    Update the policy model $\pi_{\theta}$ using \Cref{eq:visual_srft}\;
}
\KwOut{$\pi_{\theta}$}
\end{algorithm2e}

\begin{figure*}
    \centering
    \includegraphics[width=1.\linewidth]{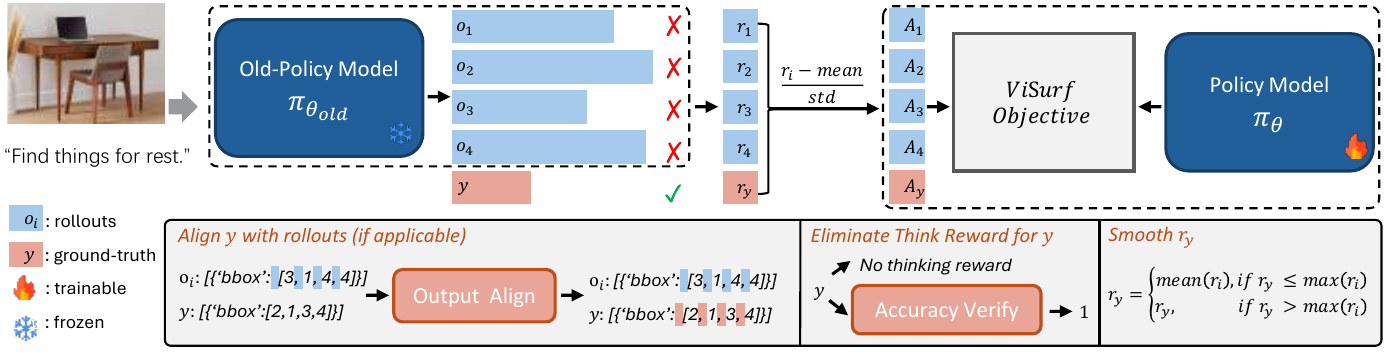}
    \caption{{\ours} Framework. Upper: The integration of external guidance $y$ with internal guidance $o_i$, which is critical when self-rollouts are unsuccessful. Bottom: Three reward control strategies designed to regulate $y$, thereby preventing entropy collapse.}
    \label{fig:architecture}
\end{figure*}

\textbf{Gradient Analysis of {\ours}.} In our practical implementation, we adopt a single-sampling, single-updating paradigm, which effectively bypasses the need for clipping operations. Therefore, the gradient of \Cref{eq:visual_srft} can be derived using approximation $\pi_{\theta} \approx \pi_{\theta_{old}}$ and log-derivative trick:

\begin{equation}
\footnotesize
\label{eq:delta_visualsrft}
\begin{aligned}
\nabla_{\theta} \mathcal{L}_{\mathrm{{\ours}}} & (\theta)
= 
-\mathbb{E}_{\substack{(v,t)\sim\mathcal{D}_{\mathrm{input}} \\
                          \textcolor{rftcolor}{\{o_j\}_{j=1}^G \sim \pi_{\theta_{\mathrm{old}}}} \\
                          \textcolor{sftcolor}{y \sim \mathcal{D}_{\text{label}}}}} \\
&\Bigg[ \frac{1}{G+1}\Bigg( 
    \sum_{j=1}^G \textcolor{rftcolor}{\hat A_j} \, \nabla_{\theta}\log\pi_{\theta}(o_j \mid v,t) \\
&\quad + \textcolor{sftcolor}{\hat A_y} \, \nabla_{\theta}\log\pi_{\theta}(y \mid v,t)
\Bigg) \Bigg]_{\,\theta \approx \theta_{\mathrm{old}}}.
\end{aligned}
\end{equation}

\subsection{Reward Control for Ground-Truth Label}
\label{sec:reward_control}
The advantage $\hat{A}y$ for the ground-truth label $y$ is always positive till now, as correct labels inherently receive higher rewards. However, this static setup is often sub-optimal; it can lead to reward hacking and suppresses the relative advantage $\hat{A}_j$ of self-generated rollouts, even when the policy has already produced correct trace and answers. Furthermore, maintaining the approximation $\pi_{\theta} \approx \pi_{\theta_{old}}$ requires ensuring that ground-truth data distributions remain aligned with self-generated rollouts. To mitigate these issues, we propose three reward control strategies:

\textbf{Aligning Ground-truth Labels with Rollouts Preference. }To ensure compatibility between the ground-truth annotations and the outputs generated by $\pi_{\theta}$, we reformat the ground-truth data to match the model’s preferred stylistic patterns. For instance, we adjust the whitespace in JSON-like structures from \{`bbox':[x1,y1,x2,y2]\} to \{`bbox':\colorbox{sftcolor}{\vphantom{x}}[x1,\colorbox{sftcolor}{\vphantom{x}}y1,\colorbox{sftcolor}{\vphantom{x}}x2,\colorbox{sftcolor}{\vphantom{x}}y2]\} (i.e., adding space after the punctuation), because minor variations in punctuation can lead to significantly different tokenizations. This alignment minimizes the distribution shift between $\pi_{\theta}$ and $\pi_{\theta_{old}}$, thereby upholding the core assumptions $\pi_{\theta} \approx \pi_{\theta_{old}}$.

\textbf{Eliminating Thinking Reward for Ground-truth Labels.} Since the ground-truth labels lack annotated reasoning path, we assign a reasoning format score of zero to them when reasoning is needed. This ensures that the model learns to generate its own reasoning traces through self-rollouts without being biased or penalized by the absence of external reasoning annotations in the ground-truth data.

\textbf{Smoothing the Reward for Ground-truth Labels.} Prior to advantage estimation, we compare the maximum reward among generated rollouts, $\max\{\{\text{r}(o_j)\}_{j=1}^G\}$, against the ground-truth reward $\text{r}(y)$. If $\max\{\{\text{r}(o_j)\}_{j=1}^G\} \geq \text{r}(y)$, it indicates the policy model $\pi_{\theta}$ has already produced a high-quality output without external guidance. In this case, we set $\text{r}(y)=\text{mean}\{\{\text{r}(o_j)\}_{j=1}^G\}$. This smoothing ensures that the advantage for the ground-truth, $\hat A_y$, becomes zero (as per \Cref{eq:adv_y}), eliminating the external supervision signal.

\subsection{Optimization Analysis During Training}
\label{sec:optimization_analysis}

To better analysis the optimization step, we reformulate the gradient \Cref{eq:delta_visualsrft} as following:

\begin{equation}
\footnotesize
\label{eq:delta_visualsrft_2}
\begin{aligned}
\nabla_{\theta} & \mathcal{L}_{\mathrm{{\ours}}}(\theta) = \\[1mm]
& \coloredunderbrace{rftcolor}{%
   -\,\mathbb{E}_{\substack{(v,t)\sim\mathcal{D}_{\mathrm{input}} \\
         \textcolor{rftcolor}{\{o_j\}_{j=1}^G \sim \pi_{\theta_{\mathrm{old}}}}}} 
   \Biggl[ 
     \frac{1}{G+1}\sum_{j=1}^G \textcolor{rftcolor}{\hat A_j} \,\nabla_{\theta}\log\pi_{\theta}(o_j\mid v,t)
   \Biggr]_{\theta\approx\theta_{\mathrm{old}}}
}{RLVR Term} \\[1mm]
& \coloredunderbrace{sftcolor}{%
   -\,\mathbb{E}_{\substack{(v,t)\sim\mathcal{D}_{\mathrm{input}} \\
        \textcolor{sftcolor}{y \sim \mathcal{D}_{\text{label}}}}} 
   \Biggl[
     \frac{1}{G+1}\textcolor{sftcolor}{\hat A_y} \,\nabla_{\theta}\log\pi_{\theta}(y\mid v,t)
   \Biggr]_{\theta\approx\theta_{\mathrm{old}}}
}{SFT Term} .
\end{aligned}
\end{equation}

The RLVR term in \Cref{eq:delta_visualsrft_2} is structurally identical to the standard RLVR gradient in \Cref{eq:delta_rft}, differing only in its scaling coefficient ($\frac{1}{G+1} \hat A_j$ vs. $\frac{1}{G} \hat A_j$). 
Similarly, the SFT term in \Cref{eq:delta_visualsrft_2} resembles the SFT gradient from \Cref{eq:delta_sft}, with two key distinctions: (i)  the coefficient is weighted by $\frac{1}{G+1} \hat A_y$ instead of 1, and (ii) the use of the approximation $\pi_{\theta} \approx \pi_{\theta_{old}}$. 
Crucially, \Cref{eq:delta_visualsrft} integrates both the external guidance from SFT and the internal guidance from RLVR.

\setlength{\tabcolsep}{8.5pt}{
\begin{table*}[t]
    \caption{Comparison on different benchmarks in different domains under different training paradigms.}
    \footnotesize
    \centering
        \begin{tabular}{c|cccccccc|c}
            \toprule
            \multirow{4}{*}{\textbf{Method}} & \multicolumn{2}{c}{Non-Object} & \multicolumn{2}{c}{Segmentation} & GUI & Anomaly & Medical:Skin & Math & \multirow{4}{*}{\textbf{Avg}} \\
            & \multicolumn{2}{c}{\textbf{gRefCOCO}} & \multicolumn{2}{c}{\textbf{ReasonSeg}} & \textbf{OmniACT} & \textbf{RealIAD} & \textbf{ISIC2018} & \textbf{MathVista} \\
            & \multicolumn{2}{c}{val} & val &test & test & subset & test & test-mini \\
            \cmidrule{2-9}
            & gIoU & N-Acc & gIoU & gIoU & Acc & ROC\_AUC & Bbox\_Acc & Acc\\
            \midrule
            Baseline & 33.4 & ~~1.8 & 56.9 & 52.1 & 60.4 & 50.1 & 78.8 & 68.2 & 50.2 \\ 
            SFT & 41.6 & ~~3.3 & 63.8 & 60.3 & 55.4 & 65.5 & 91.7 & 68.3 & 56.2 \\ 
            RLVR & 42.8 & ~~0.0 & 66.0 & 63.2 & 65.5 & 50.0 & 90.3 & 71.2 & 56.1 \\ 
            SFT \textrightarrow ~RLVR & 65.0 & 52.1 & 57.2 & 55.2 & 64.5 & 66.9 & 93.6 & 68.5 & 65.4 \\ 
            \cellcolor[HTML]{efefef}{{\ours}} & \cellcolor[HTML]{efefef}{\textbf{66.6}} & \cellcolor[HTML]{efefef}{\textbf{57.1}} & \cellcolor[HTML]{efefef}{\textbf{66.5}} & \cellcolor[HTML]{efefef}{\textbf{65.0}} & \cellcolor[HTML]{efefef}{\textbf{65.6}} & \cellcolor[HTML]{efefef}{\textbf{69.3}} & \cellcolor[HTML]{efefef}{\textbf{94.7}} & \cellcolor[HTML]{efefef}{\textbf{71.6}} & \cellcolor[HTML]{efefef}{\textbf{69.6}} \\ 
            \bottomrule
        \end{tabular}
    \label{tab:main_result}
\end{table*}}

Building on the reward control strategy in \Cref{sec:reward_control}, we analyze the dynamics of the terms in \Cref{eq:delta_visualsrft_2} throughout training.
As defined by \Cref{eq:adv_rollouts,eq:adv_y}, the advantages $\hat A_j$ (for rollouts) and $\hat A_y$ (for the ground-truth) govern the balance between the RLVR and SFT terms.
This balance is self-adaptive. When the policy fails to generate high-quality rollouts, $\hat A_j$ decreases (potentially becoming negative), while $\hat A_y$ remains high. Consequently, the SFT term dominates the policy update, providing strong external guidance from the ground-truth label. Conversely, when the policy successfully generates desirable rollouts, our reward control mechanism sets $\hat A_y \approx 0$, causing the optimization to be dominated entirely by the RLVR term. This automatic shifting between learning modes is a core feature of the single-stage {\ours} paradigm.

\textbf{Upper Bound Analysis.} As mentioned above, our {\ours} is particularly beneficial when old policy model $\pi_{\theta_{old}}$ cannot generate correct rollouts. When the old policy model $\pi_{\theta_{old}}$ already achieves desirable rollouts, the SFT Term in \Cref{eq:delta_visualsrft_2} equals to zero, thus the upper bound of {\ours} is the RLVR. However, when the policy model cannot generate desirable rollouts, the upper bound is better than using either SFT or RLVR alone.

\section{Experiments}
\Cref{sec:expri_setting} details the experimental settings. We validate {\ours} across diverse domains in \Cref{sec:comparison_sft_rl_srft}, followed by an ablation of the reward control design in \Cref{sec:control_reward_experi}. Finally, \Cref{sec:in_depth_analysis} provides a in-depth analysis of {\ours}.

\subsection{Experimental Settings}
\label{sec:expri_setting}
We verify {\ours} on benchmarks across several domains, including Non-Object Segmentation (e.g., gRefCOCO \citep{liu2023gres}), Reasoning Segmentation (e.g., ReasonSeg \citep{lai2024lisa}), GUI Grounding (e.g., OmniACT \citep{kapoor2024omniact}), Industrial Anomaly Detection (e.g., RealIAD \citep{wang2024real}), Medical Imaging (e.g., ISIC2018 \citep{codella2019skin}), and Mathematical Reasoning (e.g., MathVista \citep{lu2023mathvista}).
Details of data split and evaluation protocols are provided in the Appendix \ref{sec:detailed_experiment_settings}. 

\textbf{Implemention Details.} We instantialize {\ours} algorithm with Qwen2.5VL-7B \citep{bai2025qwen25vl} and adopt SAM2 \citep{ravi2024sam2} if needed. We employ a constant learning rate of 1e-6 for all methods, with a batch size of 32 for SFT and 16 for RLVR and {\ours}. We employ same training steps for fair comparison. For MathVista, the reward function consists of format and accuracy rewards. For other tasks, we adopt the rewards from VisionReasoner \citep{liu2025visionreasoner}, which include format accuracy, point accuracy, and bounding box accuracy rewards, etc.

\setlength{\tabcolsep}{15pt}{
\begin{table}[t]
    \caption{Comparison on VQA under different training paradigms.}
    \footnotesize
    \centering
        \begin{tabular}{c|cc}
            \toprule
            \textbf{Method} & \textbf{ChartQA} & \textbf{DocVQA\_val} \\
            \midrule
            Baseline & 83.8 & 94.9 \\
            SFT & 80.8 & 89.6 \\
            RLVR & 86.7 & \textbf{95.0} \\
            SFT \textrightarrow ~RLVR & 85.0 & 92.9\\
            \cellcolor[HTML]{efefef}{{\ours}} & \cellcolor[HTML]{efefef}{\textbf{87.4}} & \cellcolor[HTML]{efefef}{\textbf{95.0}} \\
            \bottomrule
        \end{tabular}
    \label{tab:visual_qa}
\end{table}}

\subsection{Comparison of Different Training Paradigms}
\label{sec:comparison_sft_rl_srft}
We compare different post-training paradigms and verify the effectiveness of {\ours} in various domains.

\textbf{Main Results.} A comparative analysis of various post-training paradigms is summarized in Table \ref{tab:main_result}. Empirical evaluations demonstrate that {\ours} consistently surpasses existing methodologies across all evaluated domains, achieving a substantial average relative improvement of 38.6\% over the baseline. This performance gain is most pronounced in challenging domains such as Non-Object and Anomaly, where the baseline initially struggles. This suggests that {\ours} is particularly effective at expanding a model's capabilities in areas that exceed its inherent knowledge base. Conversely, in domains where the baseline already exhibits high proficiency, we observe more marginal incremental improvements.
Our analysis also reveals that standard Supervised Fine-Tuning (SFT) leads to performance degradation on OmniACT, a phenomenon likely stemming from overfitting during the baseline’s original pre-training phase. In contrast, both RLVR and {\ours} successfully preserve the baseline’s foundational performance. Notably, in RealIAD and gRefCOCO (non-object detection), the pure RLVR approach actually underperforms the original model. We attribute this to the high frequency of incorrect self-generated rollouts, which introduce noise and hinder effective model optimizationm, a bottleneck that {\ours} effectively mitigates.

\textbf{Catastrophic Forgetting.} We evaluate the performance of ChartQA \citep{masry2022chartqa} and DocVQA \citep{mathew2021docvqa} after fine-tuning without VQA data. As illustrated in \Cref{tab:visual_qa}, VQA performance exhibits notable variation across different training paradigms. Both RLVR and {\ours} demonstrate robustness against catastrophic forgetting. In contrast, SFT and SFT \textrightarrow ~RLVR suffer from performance degradation, which is attributable to catastrophic forgetting.

\setlength{\tabcolsep}{17pt}{
\begin{table}[t]
    \caption{Employ {\ours} on Qwen2VL-7B.}
    \footnotesize
    \centering
        \begin{tabular}{c|cc}
            \toprule
            \multirow{3}{*}{\textbf{Method}} & \textbf{RealIAD} & \textbf{ISIC2018} \\
            & subset &  test \\
            \cmidrule{2-3}
            & ROC\_AUC & Bbox\_Acc \\
            \midrule
            Baseline & 60.0 & 51.8 \\
            SFT & 56.7 & 94.2 \\
            RLVR & 57.1 & 90.5  \\
            SFT \textrightarrow ~RLVR & 67.5 & 94.6 \\
            \cellcolor[HTML]{efefef}{{\ours}} & \cellcolor[HTML]{efefef}{\textbf{76.0}} &  \cellcolor[HTML]{efefef}{\textbf{95.4}} \\
            \bottomrule
        \end{tabular}
    \label{tab:other_models}
\end{table}}

\textbf{{\ours} on Other Models.} We apply {\ours} to the Qwen2VL-7B \citep{wang2024qwen2vl}. As shown in \Cref{tab:other_models}, our method consistently outperforms its counterparts. 

\setlength{\tabcolsep}{1pt}{
\begin{table}[t]
    \caption{Ablation of Reward Control Strategy in \Cref{sec:reward_control}. The first row is the non-trained baseline. `Align': Aligning ground-truth labels with rollouts; `Eliminate': Eliminating thinking format reward for groud-truth labels; `Smooth': Smoothing accuracy reward for ground-truth labels; `-': not applicable.}
    \footnotesize
    \centering
        \begin{tabular}{ccc|cc|c|c}
            \toprule
            \multirow{3}{*}{\textbf{Align}} & \multirow{3}{*}{\textbf{Eliminate}} & \multirow{3}{*}{\textbf{Smooth}} &  \multicolumn{2}{c}{\textbf{gRefCOCO}} \vline &  \textbf{ReasonSeg} &  \textbf{MathVista}\\
            &  &  &  \multicolumn{2}{c}{val} \vline & val & testmini \\
            \cmidrule{4-7} 
            & & & gIoU & N-Acc & gIoU & Acc \\
            \midrule
            --  & -- & -- & 33.4 & 1.8 & 56.9 & 68.2 \\
            \ding{55} & \checkmark & \checkmark & 59.0 & 40.2 & 63.6 & ---\\
            \checkmark & \ding{55} & \checkmark & \textbf{72.9} & \textbf{74.1} & 58.2 & 67.1\\
            \checkmark & \checkmark & \ding{55} & 61.0 & 45.7 & 62.7 & 66.8\\
            
            \checkmark & \checkmark & \checkmark & 66.6 & 57.1 & \textbf{66.5} & \textbf{71.6}\\
            
            \bottomrule
        \end{tabular}
    \label{tab:ablation_reward_control}
\end{table}}

\subsection{Ablation of Reward Control}  
\label{sec:control_reward_experi}

\Cref{tab:ablation_reward_control} presents an ablation study of the reward control strategy for ground-truth labels, detailed in \Cref{sec:reward_control}.

\textbf{Aligning Ground-truth Labels with Rollouts Preference.} The empirical results underscore the critical importance of this strategy, as its ablation leads to consistent performance degradation across multiple datasets. This observation provides strong empirical validation for the theoretical requirement of $ \pi_{\theta} \approx \pi_{\theta_{old}} $ presented in \Cref{eq:delta_visualsrft_2}.

\textbf{Eliminating Thinking Reward for Ground-truth Labels.} The results indicate that the reasoning strategy is critical for tasks requiring complex inference, such as those in ReasonSeg and MathVista, as it encourages the model to generate a reasoning process prior to delivering the final answer. Conversely, for the gRefCOCO dataset, where queries are typically limited to simple class types (e.g., "human") and basic references (e.g., "woman on the right"), omitting the reasoning trace yields superior performance. This suggests that the necessity of explicit reasoning is contingent upon the complexity of the underlying task.

\textbf{Smoothing the Reward for Ground-truth Labels.} The performance decline observed across datasets following the ablation of reward smoothing underscores its necessity. Concurrently, the results suggest that the SFT Term in \Cref{eq:delta_visualsrft_2} becomes superfluous when the model's self-rollouts already achieve a higher-quality solution.

\subsection{In-depth Analysis}
\label{sec:in_depth_analysis}
To facilitate a deeper understanding of {\ours}, we provide a detailed analysis of its behavior and properties.

\begin{figure}[t]
    \centering
    \includegraphics[width=0.95\linewidth]{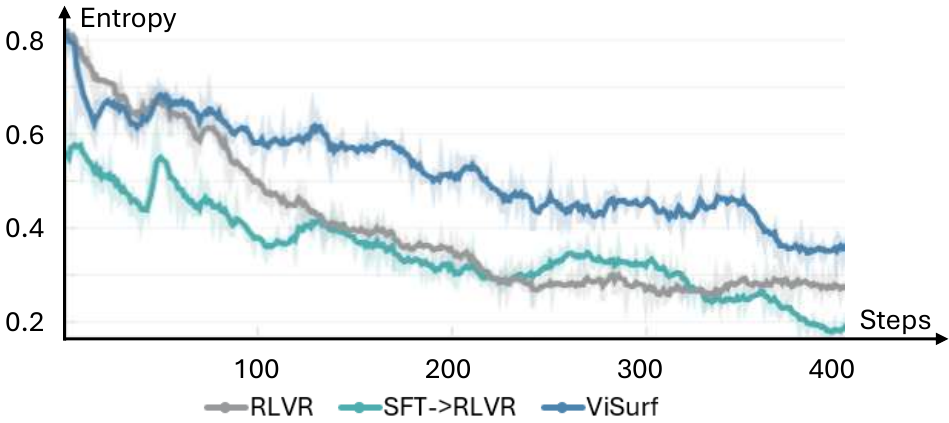}
    \caption{Entropy Analysis of RLVR, SFT \textrightarrow RLVR and {\ours}. {\ours} exhibits an initial drop, then converges slowly.}
    \label{fig:entropy}
\end{figure}

\textbf{Entropy Analysis During Training.} \Cref{fig:entropy} compares the entropy of RLVR, SFT \textrightarrow ~RLVR and {\ours}. A higher entropy indicates greater exploratory behavior, while lower entropy signifies convergence toward certainty. We observe that {\ours} exhibits an initial entropy drop, indicating the model is fitting the external guidance. Subsequently, {\ours} converges at a slower rate than others, thereby effectively avoiding entropy collapse.

\begin{figure}
    \centering
    \includegraphics[width=0.95\linewidth]{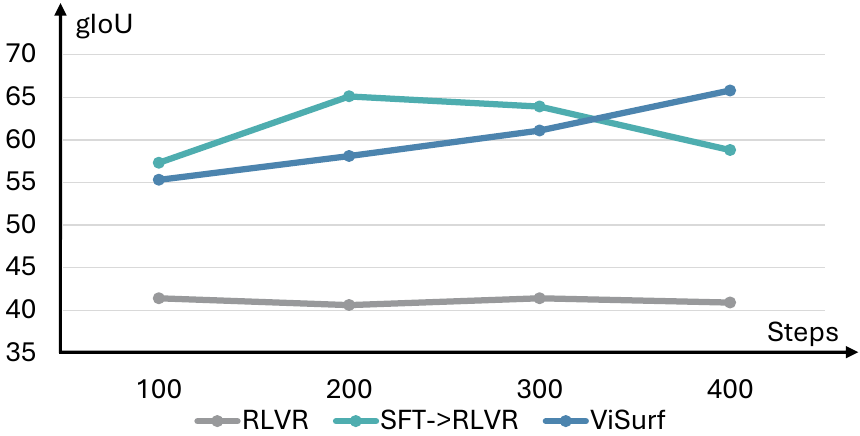}
    \caption{Performance on gRefCOCO in different training steps. {\ours} demonstrates greater stability as training proceeds.}
    \label{fig:train_stability}
\end{figure}

\textbf{Training Stability.} \Cref{fig:train_stability} demonstrates that models trained with our method exhibit greater stability than those trained with pure RLVR and SFT \textrightarrow ~RLVR, as the performance of others decline with longer training. This observation confirms the effectiveness of our approach, indicating that the introduced external guidance acts as a constraint, which stabilizes the training process.

\textbf{Boundary Analysis.} As shown in \Cref{tab:main_result}, the performance gain of {\ours} is related to the baseline model's performance. When the baseline performs poorly (e.g., below 50\%), indicating its inadequacy for the task, our method yields a substantial improvement. Conversely, when the baseline already achieves high performance (e.g., above 50\%), signifying a strong starting point, the upper bound of our method aligns with that of RLVR alone. This observation corroborates our theoretical analysis in \Cref{sec:optimization_analysis}.

\setlength{\tabcolsep}{8pt}{
\begin{table}[t]
    \caption{Comparison of different prompt design. {\ours} achieves satisfying results even without detailed formatting prompt.}
    \footnotesize
    \centering
        \begin{tabular}{cc|cc}
            \toprule
            \multirow{3}{*}{} & \multirow{3}{*}{\textbf{Detailed Prompt}} & \multicolumn{2}{c}{\textbf{ReasonSeg}} \\
            & & val (gIoU) & test (gIoU) \\
            \midrule
            \multirow{2}{*}{RLVR} & \ding{55}  & ~~0.0 & ~~0.0 \\
                                 &  \checkmark & 66.0 & 63.2\\
            \midrule
            \multirow{2}{*}{{\ours}}  & \ding{55}  & 62.3 & 57.8 \\
                                     & \checkmark  & \textbf{66.4} & \textbf{65.0} \\
            \bottomrule
        \end{tabular}
    \label{tab:prompt_burden}
\end{table}}

\textbf{Reduce the Burden of Prompt Design.} The RLVR paradigm relies heavily on explicit instructions to guide the model toward generating rollouts in a specific format, such as \textit{output with format 'point\_2d': [2, 3]}. In contrast, {\ours} incorporates external guidance with desired format, thereby reducing the dependency on manual prompt engineering. \Cref{tab:prompt_burden} compares performance with and without detailed prompts (\textit{prompts difference are shown in the supplementary materials}), demonstrating that our approach achieves consistent gains in both settings, whereas RLVR fails without explicit formatting instructions.

\setlength{\tabcolsep}{8pt}{
\begin{table}[t]
    \caption{Comparison of training cost of different training paradigms with same batch size. Time for two-stage SFT \textrightarrow ~RLVR is estimated as the addition of SFT and RLVR. }
    \footnotesize
    \centering
        \begin{tabular}{c|cc}
            \toprule
            \textbf{Method} & \textbf{Mem / GPU (G) \textdownarrow} & \textbf{Time / Step (s) \textdownarrow} \\
            \midrule
            SFT & 97.7 & ~~\textbf{9.0} \\
            RLVR & \textbf{81.8} & 22.7\\
            SFT \textrightarrow ~RLVR & 97.9 & 31.7\\
            \cellcolor[HTML]{efefef}{{\ours}} & \cellcolor[HTML]{efefef}{\textbf{81.8}} & \cellcolor[HTML]{efefef}{22.9} \\
            \bottomrule
        \end{tabular}
    \label{tab:training_cost}
\end{table}}

\textbf{Training Cost.} We conducted a comparative analysis of the per-step training for different fine-tuning paradigms. Each method is implemented using well-established frameworks: DeepSpeed \citep{deepspeed} and TRL \citep{trl} for SFT, and VeRL \citep{verl} for RLVR and {\ours}. The results indicate that while RLVR and {\ours} offer greater memory efficiency, they increase computational time, attributable to the overhead of generating rollouts.

\subsection{Comparison with State-of-The-Arts}
\label{sec:sota_methods}
We compare {\ours} against state-of-the-art (SoTA) models on two visual perception tasks: gRefCOCO and ReasonSeg. We compare LISA \citep{lai2024lisa}, GSVA \citep{xia2024gsva}, SAM4MLLM \citep{chen2024sam4mllm}, SegZero \citep{liu2025seg}, VisionReasoner \citep{liu2025visionreasoner}. As shown in \Cref{{tab:compare_sotas}}, {\ours} achieves the SoTA performance.

\setlength{\tabcolsep}{2pt}{
\begin{table}[t]
    \caption{Comparison with SoTAs. `-' means not available.}
    \footnotesize
    \centering
        \begin{tabular}{l|cccc}
            \toprule
            \multirow{3}{*}{\textbf{Method}} & \multicolumn{2}{c}{\textbf{gRefCOCO}} & \multicolumn{2}{c}{\textbf{ReasonSeg}} \\
            & \multicolumn{2}{c}{val} & val & test \\
            \cmidrule{2-5}
            & gIoU & N-Acc & gIoU & gIoU \\
            \midrule
            LISA-7B & 61.6 & 54.7 & 53.6 & 48.7  \\
            GSVA-7B & 66.5 & 62.4 & - & -  \\
            SAM4MLLM-7B & 69.0 & 63.0 & 46.7 & - \\
            Qwen2.5VL-7B + SAM2 & 41.6 & ~~3.3 & 56.9 & 52.1 \\
            SegZero-7B & - & - & 62.6 & 57.5 \\
            VisionReasoner-7B & 41.5 & ~~0.0 & 66.3 & 63.6  \\
            \cellcolor[HTML]{efefef}{{\ours} (Qwen2.5VL-7B + SAM2)} & \cellcolor[HTML]{efefef}{\textbf{72.9}} & \cellcolor[HTML]{efefef}{\textbf{74.1}} & \cellcolor[HTML]{efefef}{\textbf{66.5}} & \cellcolor[HTML]{efefef}{\textbf{65.0}}  \\
            \bottomrule
        \end{tabular}
    \label{tab:compare_sotas}
\end{table}}

\begin{figure}[t]
    \centering
    \includegraphics[width=.98\linewidth]{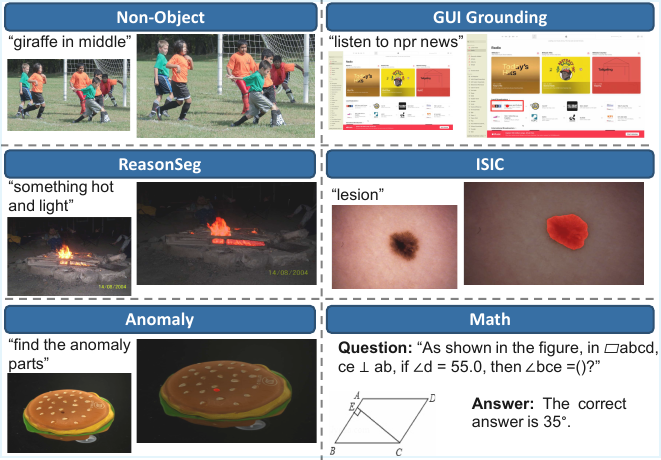}
    \caption{Visualization of {\ours} on various tasks.}
    \label{fig:visualization}
\end{figure}

\subsection{Qualitative Results}
\label{sec:qualitative}

Qualitative results on various tasks are presented in \Cref{fig:visualization}. The results demonstrate that models trained with {\ours} successfully solves multiple visual perception tasks.

\section{Conclusion}

We present {\ours}, a unified post-training paradigm designed to bridge the gap between Supervised Fine-Tuning (SFT) and Reinforcement Learning from Verifiable Rewards (RLVR). Driven by a rigorous theoretical analysis of their respective objectives and gradient dynamics, {\ours} integrates the strengths of both approaches into a single-stage framework. Our implementation strategically interleaves ground-truth labels with model-generated rollouts, utilizing three novel reward control strategies to maintain training stability. Empirical evaluations across diverse benchmarks demonstrate that {\ours} consistently outperforms standalone SFT, RLVR, and the traditional sequential SFT $\rightarrow$ RLVR pipeline. Our in-depth analysis provides deep insights into the framework's behavior, effectively corroborating our theoretical foundations.

\section*{Acknowledgements}
This work was supported in part by the Research Grants Council under the Areas of Excellence scheme grant AoE/E-601/22-R.

\section*{Impact Statement}
This paper presents work whose goal is to advance the field of machine learning. There are many potential societal consequences of our work, none of which we feel must be specifically highlighted here.


\bibliography{example_paper}
\bibliographystyle{icml2026}

\appendix
\clearpage

\section{Limitations}
The current reinforcement learning codebase relies on VeRL~\citep{verl}, which mainly support the Qwen-VL series \citep{bai2025qwen25vl,wang2024qwen2vl}. Extending support to additional architectures is left for future work as framework compatibility improves.

\section{Discussion and Future Work}
The principal insight of {\ours} is the effective combination of RLVR's internal reinforcement and the external guidance of SFT. Although the ground-truth labels in this work are limited to final answers, our {\ours} paradigm is inherently compatible to incorporate explicit reasoning traces. The flexibility also ensures compatibility with advanced techniques like knowledge distillation, where reasoning traces from larger models could be directly incorporated. We anticipate that this work will provide a foundation for future research in LVLMs' post-training.

\section{Detailed Experimental Settings}
\label{sec:detailed_experiment_settings}
\textbf{Non-Object Segmentation.} The gRefCOCO \citep{liu2023gres} includes queries that do not contain corresponding objects. The evaluation metrics are gIoU and N-Acc. We use Multi-objects-7K  \cite{liu2025visionreasoner} plus with 200 non-object data for training.

\textbf{Reasoning Segmentation.} The ReasonSeg \citep{lai2024lisa} includes test samples that need reasoning for correct segmentation. It has 200 validation images and 779 test images. The evaluation metric is gIoU. We use Multi-objects-7K \cite{liu2025visionreasoner} proposed in VisionReasoner \citep{liu2025visionreasoner} for training.

\textbf{GUI Grounding.} The OmniACT \citep{kapoor2024omniact} is a GUI grounding task for Desktop and Web. We randomly collect 6,101 samples in training split and verify on the test split. The accuracy is calculated as whether the predict point correctly locates inside the interest region.

\textbf{Anomaly Detection.} The RealIAD \citep{wang2024real} includes real-world, multi-view industrial anomaly. We derive 3,292 training samples and 2,736 test samples, ensuring the two sets are disjoint. We calculated the ROC\_AUC. 

\textbf{Medical Image: Skin.} The task one of ISIC2018 \citep{codella2019skin,isic2018} is lesion segmentation. It includes 2,594 training samples and 1,000 test samples. We measure the bbox\_acc metric, which computes the ratio of predicted bboxes whose IoU with the ground truth exceeds 0.5.

\textbf{MathVista.} The MathVista-testmini \citep{lu2023mathvista} includes 1,000 diverse mathematical and visual tasks. We gather around 10k training data from WeMath \citep{qiao2024we}, MathVision \citep{wang2024measuring}, Polymath \citep{gupta2024polymath}, SceMQA \citep{liang2024scemqa}, Geometry3K \citep{lu2021inter}.

\section{Illustration of Prompt Design}
\label{sec:prompt_design}

Prompt~\ref{with_format_instruction} shows prompt with detailed format instruction, where we provide desired output format for model. Prompt~\ref{without_format_instruction} shows prompt without detailed format instruction, where we simply write `answer here' between answer tags.

\begin{promptbox}{\textbf{\textcolor{red!30!white}{without}} detailed format instruction}{without_format_instruction}
``Please find `\{\{\textit{Question}\}\}\}' with bboxs and points.'' \\
``Compare the difference between object(s) and find the most closely matched object(s).'' \\
``Output the thinking process in \textless{}think\textgreater{} \textless{}/think\textgreater{} and final answer in \textless{}answer\textgreater{} \textless{}/answer\textgreater{} tags.'' \\
``Output the bbox(es) and point(s) inside the interested object(s) in JSON format.''

\begin{verbatim}
i.e. <think> 
thinking process here 
</think>
<answer> 
answer here 
</answer>
\end{verbatim}
\end{promptbox}

\begin{promptbox}{\textbf{\textcolor{red!30!white}{with}} detailed format instruction}{with_format_instruction}
``Please find `\{\{\textit{Question}\}\}\}' with bboxs and points.'' \\
``Compare the difference between object(s) and find the most closely matched object(s).'' \\
``Output the thinking process in \textless{}think\textgreater{} \textless{}/think\textgreater{} and final answer in \textless{}answer\textgreater{} \textless{}/answer\textgreater{} tags.'' \\
``Output the bbox(es) and point(s) inside the interested object(s) in JSON format.''

\begin{verbatim}
i.e. <think> 
thinking process here
</think>
<answer>
[{"bbox_2d": [10, 100, 200, 210], 
"point_2d": [30, 110]}, 
{"bbox_2d": [225, 296, 706, 786], 
"point_2d": [302, 410]}]
</answer>
\end{verbatim}
\end{promptbox}

\section{Additional Explanation on pure RLVR}
\label{sec:addition_explanation}

The performance of pure RLVR is heavily influenced by randomness. In rare cases, the old policy model $\pi_{\theta_{old}}$ can generate correct non-object outputs in the initial steps, thereby achieving competitive performance (see \Cref{tab:additional_grefcoco}), yet still marginally inferior to {\ours}.

\setlength{\tabcolsep}{20pt}{
\begin{table}[h]
    \caption{Comparison under different training paradigms.}
    \footnotesize
    \centering
        \begin{tabular}{c|cc}
            \toprule
            \multirow{3}{*}{\textbf{Method}} & \multicolumn{2}{c}{gRefCOCO} \\
            & \multicolumn{2}{c}{val} \\
            \cmidrule{2-3}
            & gIoU & N-Acc \\
            \midrule
            Baseline & 33.4 & ~~1.8 \\ 
            SFT & 41.6 & ~~3.3 \\
            RLVR & 42.8 & ~~0.0 \\
            RLVR (rare) & 62.9 & 49.3 \\
            SFT \textrightarrow ~RLVR & 58.6 & 38.1 \\
            \cellcolor[HTML]{efefef}{{\ours}} & \cellcolor[HTML]{efefef}{\textbf{66.6}} & \cellcolor[HTML]{efefef}{\textbf{57.1}}  \\
            \bottomrule
        \end{tabular}
    \label{tab:additional_grefcoco}
\end{table}}

\section{Non Object Training Data}
\label{sec:non_object_train}
The training data for gRefCOCO was adapted from the VisionReasoner training data \citep{liu2025visionreasoner}, from which we utilized 7k referring expression samples. To enhance the model's ability to handle cases where no target object is present, we augmented this dataset with 200 non-object examples. These negative samples were generated by providing a question that is unanswerable given the image content and training the model to output an empty list (\textless answer\textgreater []\textless  /answer\textgreater).

\section{Data Mixture in Non-object Scenarios}

We conduct a data sensitivity analysis for the non-object setting. As shown below, ViSurf’s performance consistently improves as the amount of non-object data increases. Notably, it can already detect absent objects with as few as 50 negative samples, demonstrating strong robustness even under limited annotated data.

\setlength{\tabcolsep}{10pt}{
\begin{table}[h]
\centering
\caption{Performance comparison with different training data mixture on gRefCOCO.}
\label{tab:data_sensitivity_grefcoco}
\footnotesize
\begin{tabular}{lcccc}
\toprule
Method & Obj & Non-Obj & gIoU & N-ACC \\
\midrule
Baseline & -- & -- & 33.4 & 1.8 \\
ViSurf & ~7k & ~50  & 61.5 & 44.4 \\
ViSurf & ~7k & 200 & 66.6 & 57.1 \\
ViSurf & 6.7k & 500 & \textbf{74.9} & \textbf{80.2} \\
\bottomrule
\end{tabular}
\end{table}}



\end{document}